\documentclass[twoside,twocolumn,10pt]{article}
\usepackage{fancyhdr}
\usepackage{footnote}
\usepackage{tabularx}
\usepackage{subcaption}
\usepackage{tablefootnote}
\usepackage{multicol}
\usepackage{abstract}
\usepackage{graphicx}
\usepackage{caption}
\usepackage{dblfloatfix}
\usepackage{titlesec}
\usepackage{ragged2e}  
\usepackage{amssymb}  
\usepackage{parskip}  
\usepackage{amsmath}  
\usepackage{newtxtext}  
\usepackage{booktabs}  
\usepackage{array}     
\usepackage[round]{natbib}
\usepackage{hyperref} 
\usepackage{geometry}
\usepackage{cleveref}
\DeclareMathOperator{\Cov}{Cov}
\DeclareMathOperator{\Dec}{Dec}
\DeclareMathOperator{\RA}{RA}

\newcolumntype{L}[1]{>{\raggedright\arraybackslash}p{#1}} 
\newcolumntype{C}[1]{>{\centering\arraybackslash}p{#1}}   
\newcolumntype{R}[1]{>{\raggedleft\arraybackslash}p{#1}}  

\usepackage[font=footnotesize, labelfont=bf, justification=raggedright, format=plain]{caption}

\titleformat{\section}{\large\bfseries\uppercase}{\thesection.}{1em}{}

\titleformat{\subsection}{\normalfont\bfseries\flushleft}{\thesubsection.}{1em}{}

\titleformat{\subsubsection}{\normalfont\bfseries\itshape\flushleft}{\thesubsubsection}{1em}{}

\newcommand{\keywordsname}{Keywords}

\fancypagestyle{firstpage}{
  \fancyhf{}  
  \fancyfoot[L]{\footnotesize \\\textsuperscript{*}Corresponding Author: Aleksandr Berdnikov \textless aberdnik@fields.utoronto.ca\textgreater} 
}

\title{\fontsize{16}{20}\selectfont\textbf{Sky sphere representation in language models}}
\author{
    \fontsize{12}{14}\selectfont
    Aleksandr Berdnikov\textsuperscript{*1,2}, Yevgeny Liokumovich\textsuperscript{3}\\[1ex]
    \fontsize{12}{14}\selectfont
    \textsuperscript{1}Fields Institute 
    \textsuperscript{2}Principles of Intelligence \\
    \textsuperscript{3}University of Toronto
}

\date{2026}

\begin{document}

\twocolumn[
\begin{@twocolumnfalse}
\vspace{-0.5cm}  
\raggedright

{\fontsize{11}{13}\selectfont{Research Article}}
\vspace{-0.8cm}  

\maketitle

\thispagestyle{firstpage} 
\begin{abstract}
\fontsize{11}{13}\selectfont
\vspace{1em}
\justifying
We analyze whether language models of size $\sim$100B have a representation of the night sky map that is decodable from their residual stream. We find that most of the considered open-source models do have such a representation, and it often even surfaces to the top principal components on prompts that ask questions like ``what is close to this object in the night sky''. In all but one model this representation showed significant scores in LOO testing, containing up to 65-85\% of variance ($R^2$-score) and having median angular error down to $12^\circ-21^\circ$. We verify that our representation is not a simple leak from a correlated flat representation. To our knowledge, this representation is the first example of a curved high-dimensional irreducible feature manifold. 

Codes used in the paper are published at \href{https://github.com/l3erdnik/Decodable-sky}{github.com/l3erdnik/Decodable-sky}

\vspace{-0.1cm}
\noindent\textbf{keywords}
 AI, Mechanistic Interpretability, Feature Manifold.

\end{abstract}

\vspace{1cm}
\end{@twocolumnfalse}
]
\section{Introduction}
\fontsize{12}{14}\selectfont

Mechanistic interpretability tools can be roughly divided into two groups:
linear methods, such as linear probes~\citep{probes,probingsurvey} and sparse autoencoders (SAEs)~\citep{monosemanticity,saes},
and methods that analyze activations with deep learning systems, such as natural language autoencoders (NLAs)~\citep{nla} and activation oracles~\citep{oracles}.
The downside of the second group is opaqueness.
The downside of the first is that linear methods miss important non-linear structures
present in the representation space.

Of particular interest, therefore, are \emph{feature manifolds}: low-dimensional
submanifolds of the activation space that correspond to a specific concept.
Feature manifolds offer a middle ground between the two groups of tools: they capture
non-linear structure while remaining explicit and human-inspectable.
For example, when asked ``what day is 5 days after Friday?'', LLMs represent the days
of the week (and, likewise, the months of the year) on a circle and solve this modular
arithmetic problem by rotating it~\citep{polar}. Similarly, when deciding where to break
a line in fixed-width text, a model tracks the running character count on a curved
one-dimensional manifold, which attention heads then twist to compare against the line
width~\citep{linebreak}. Both are examples of one-dimensional feature manifolds.

\begin{figure}
\centering
\includegraphics[width=0.45\textwidth]{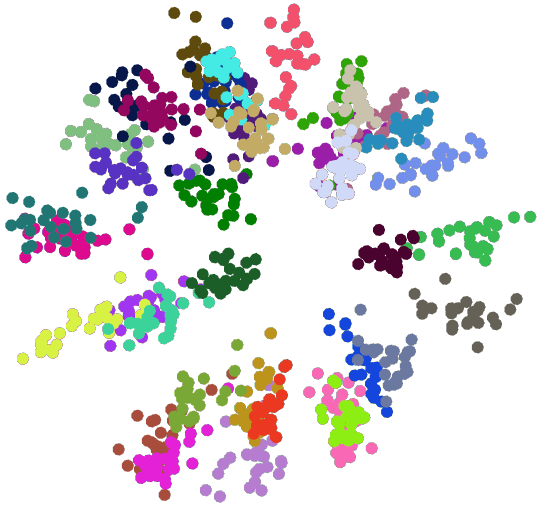}
\caption{Representation of selected 38 night sky objects in the residual stream of Mistral Large 2 LLM at layer 66. Each color corresponds to a single object point cloud.}
\label{fig:cloud}
\end{figure}

\begin{figure}[h]
\centering
\includegraphics[width=0.45\textwidth]{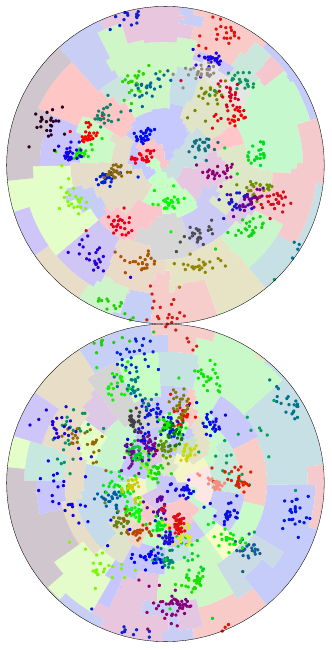}
\caption{Constellation regions in stereographic projections and corresponding point clouds from Mistral Large 2 layer 66 residual stream. Northern hemisphere on top.}
\label{fig:constellations}
\end{figure}

In this paper we demonstrate the existence of a spherical feature manifold in the 
activation space of LLMs, corresponding to the positions of stars and constellations in the night sky (Figures~\ref{fig:cloud} and~\ref{fig:constellations}). Our prompts mention the objects only by their proper names and avoid mentioning any coordinates explicitly, but rather ask what is close to the object in the sky. Thus, a spherical arrangement in activation space is likely not a trivial echo of coordinate strings seen in training, but rather a reflection of the notion of proximity for the night sky objects. Across seven open-source models of 32B--235B parameters, we find that the position of an object on the celestial sphere is linearly decodable from the top principal components of the residual stream on proximity focused prompts: in all but one model, leave-one-out
validation yields $R^2$ scores of up to 65--85\% and median angular errors down to $12^\circ$--$21^\circ$, and the representation rises into the top $\sim$8 principal components.

Previously discovered geometric representations in language models have been either flat (continents and USA states in \citep{spacetime}, color space in \citep{colors}) or one-dimensional (months and historical years in \citep{polar} and \citep{coocur}, sentence length in \citep{linebreak}]). To our knowledge, this is the first example of a high-dimensional non-flat irreducible feature manifold (that is, is not a product of 1-dimensional ones, as defined in~\citep{polar}).

\section{Methodology and Data}
\fontsize{12}{14}\selectfont

\subsection{General pipeline}
\fontsize{12}{14}\selectfont
 We feed various prompts (from a fixed list) about night sky objects into an LLM in auto-complete mode (no thinking), and extract top 128 principal components of the distribution of the residual stream vectors in every layer. We find linear regression fit (using top $n=4,8\mbox{ or }128$ PC-components) to the true coordinates of the objects on the unit sphere of the night sky. For validation on object $X$ it is left out of the data for the regression fit. We then record the accuracy of how well it predicts the unseen object $X$ across the whole dataset: the $R^2$ score (variance explained by the regression model) and median angular error. We also record the PCA components that carry the best fit (see Appendix~\ref{app:pca}).

\subsection{Input Data}
\fontsize{12}{14}\selectfont
We use 188 astronomical objects of 3 types: 85 brightest stars, all 88 official constellations and 15 arbitrary miscellaneous objects like prominent galaxies, nebula and clusters. The coordinates of the constellations are defined as their area centroids (projected back on the unit sphere) --- a choice that is somewhat arbitrary, but, we argue, inconsequential to our findings. For more, see Appendix~\ref{app:obj}. 

We feed these objects into LLMs wrapped in 25 prompts of the following type:

\begin{itemize}
    \itemsep-.75cm 
    \item the closest constellation to X is\\
    \item in star atlases X lies close to\\
    \item stargazers spot X right beside\\
\end{itemize}

where X is replaced by a given object, so the whole dataset is of size $25\times188=4700$. The prompts focus on the apparent proximity on the night sky but avoid explicit mentions of coordinates in any way. For the whole prompt set and more details see Appendix~\ref{app:prompts} 

\subsection{Models}
We used the following open-souse models in our investigation:
\begin{itemize}
    \itemsep-.75cm 
    \item Qwen3-32B\\
    \item Qwen3-235B\\
    \item gpt-oss-120b\\
    \item Llama-3.3\\
    \item Mixtral-8x22B\\
    \item Mistral Large 2\\
    \item GLM-4.5-Air\\
\end{itemize}
For more information on those see Table~\ref{tab:models}.

\section{Results}

\subsection{LLMs comparison}

\begin{figure}
\centering
\includegraphics[width=0.5\textwidth]{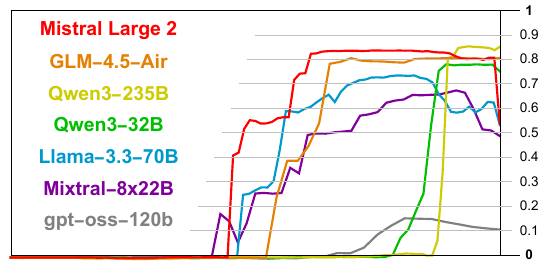}
\includegraphics[width=0.5\textwidth]{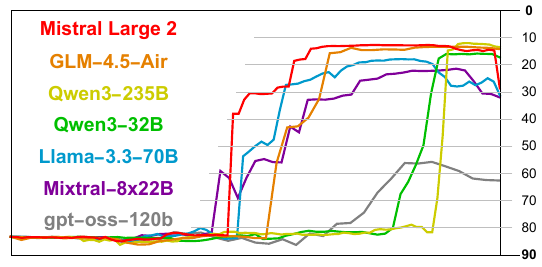}
\caption{$R^2$-score (top) and median angular error (bottom) as a function of normalized layer depth for predicting the coordinates on the night sky of an object from residual stream (top-8 PCA) using linear regression on stream data from other objects}
\label{fig:all_r2_ang}
\end{figure}

Figure~\ref{fig:all_r2_ang} shows significant decodability of the position on the celestial sphere form residual stream in most tested models. Note that the poor performance of gpt-oss in this test does not reflect poorly on its actual abilities: when asked directly for stellar coordinates, it locates over 70\% of objects to within $1^\circ$, similar to other models with thinking in our set, see Appendix~\ref{app:gpt}.

\subsection{Representation prominence in PCA}

\begin{figure}
\centering
\includegraphics[width=0.5\textwidth]{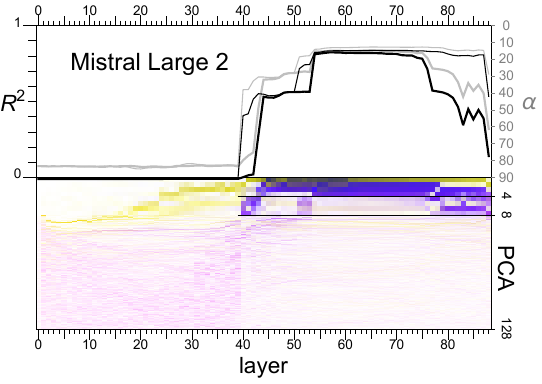}
\includegraphics[width=0.5\textwidth]{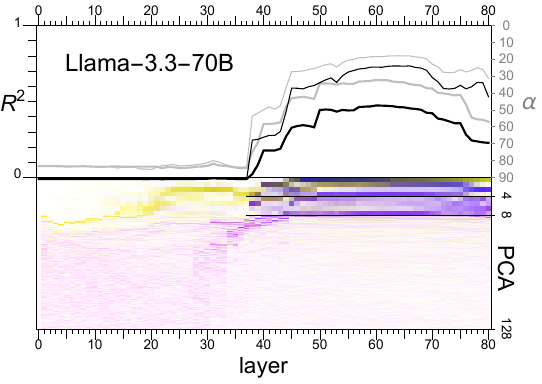}
\caption{$R^2$-score (black) and median angular error (gray) in LOO validation for sky fitting when using top-8 (thin) and top-4 (thick) PCA components. Colored charts show which PC's correlate with the optimal projection (blue\&purple) and with the object type (yellow)}
\label{fig:mist&lama}
\end{figure}

\begin{figure}
\centering
\includegraphics[width=0.5\textwidth]{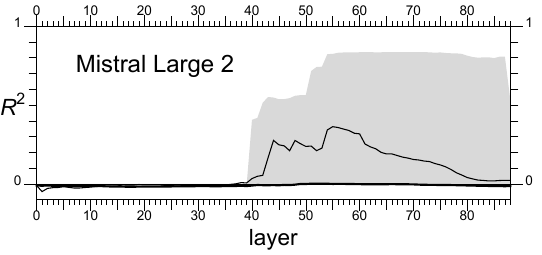}
\caption{$R^2$-score in LOO validation for sky fitting on generic prompts (see Table~\ref{tab:noloc_prompts}) compared to the top-8 fit for proximity-focused prompts (gray shade). The signal is almost 0 in top-8 components (thick) and is weak and decaying in top-32 (thin).}
\label{fig:mist_noloc}
\end{figure}

Figure~\ref{fig:mist&lama} shows that the fitting embedding of the celestial sphere for small layer depth is weak and spread along many PCA components (light and purple), but as the depth increases it strengthens and rises to top$\sim$8 PCA components (dark and blue). In many models it is only surpassed by the direction (yellow) that separates stars from constellations and leaves the sky dome mostly in PCs 2-4 (see Appendix~\ref{app:hogs}). We note that the prominence and clarity of this signal relies on our prompts being focused on night sky proximity. Figure~\ref{fig:mist_noloc} shows that on more generic prompts the signal weakens and sinks to lower PCA ranks. Similar charts for other models and their technical details are in Appendix~\ref{app:pca}.

\subsection{Competing representations}

\begin{figure}
\centering
\includegraphics[width=0.5\textwidth]{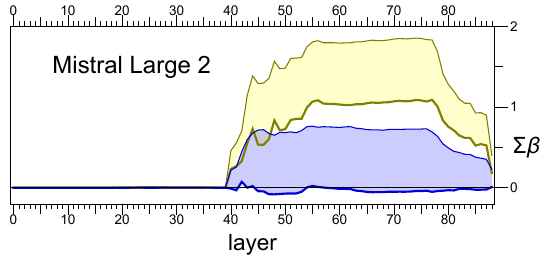}
\caption{Weights of various dimensions of a spherical (yellow) and flat 2D (blue) models in RDM analysis of top-8 PCA space. A particular pair of dimensions, that are highly correlated between the two models, makes their attribution ambiguous. If they are excluded from both models (bottom thick lines), the spherical model stays strong, but the 2D map drops to 0 and below: its discontinuous ``longitudal'' coordinate is not observed on its own.}
\label{fig:mist_2d}
\end{figure}

The unit sphere representation of the night sky in not the only possible one. In~\citep{spacetime} the model used for the continents and USA states was not globe-based, but a flat map with altitude and latitude as coordinates. Similarly, the sky may be represented as a flat 2D-chart instead of as a unit sphere. Separating the two is not entirely straightforward, since some of their dimensions are highly correlated. We use a modified RDM-analysis by measuring how much the metric on test objects (that we observe in the residual stream) is attributable to each model dimension separately. We find that the distinctive discontinuous dimension of the flat 2D chart comes with 0 (or even $<0$) coefficient in top-8 PCA space (e.g. Figure~\ref{fig:mist_2d}), which excludes the possibility of the flat chart representation there (see Appendix~\ref{app:other_maps_flat}).

Another alternative representation conceivable for the stars is a genuine 3D-representation. However, since distances to the stars alone span $\sim 3$ orders of magnitude (in our set), the radial component, if present, would have to be non-linear. We analyze the radial component, but don't find a significant (or even consistent across models) correlation with the distance to the star (Appendix~\ref{app:other_maps_rad}). This rules out our representation being secretly a 3D Euclidean one (even with adjusted, but still clear radial dependence). We find, however, that the radial component \textit{is} influenced by other factors, most notably, the type of the object (constellations are placed further, stars are placed closer), and, to a lesser extent, by its frequency in texts and, possibly, brightness (Figure~\ref{fig:radcor}).

\section{Open questions}
We only analyzed the emergence of the representation in the residual stream. That leaves several aspects unclear. Is the mechanism that constructs the representation more interpretable than pure memorization? Are the directions of the representation tied to meaningful unembeddings? Given that the models know celestial coordinates way better, than the accuracy of our representation (see Figure~\ref{fig:actual}), what is its role for the models? Our representation was extracted with proximity-focused prompts, reasoning that they could fish out ``textual proximity'', that, hopefully, is heavily influenced by proximity on the night sky. Similar logic has been used in~\citep{coocur} to explain geometric structures in activation spaces by the metric derived from co-occurence in the training data. How well is our example explained by this paradigm?

\section{Statement on AI use}
AI agent Claude Opus 4.8 was used to code and analyze experiments.

\section{Acknowledgements}
This work would not be possible without cluster access
generously offered by Cursor (https://cursor.com/).

We are grateful to Arul Shankar and Jacob Tsimerman for valuable discussions
and encouragement. 

A.B.'s research was supported by Principles of Intelligence (https://princint.ai/).

\bibliography{sky.bib}
\bibliographystyle{plainnat}

\onecolumn

\appendix
\section{Models}
\label{app:models}

\begin{table}[h]
\centering
\resizebox{\columnwidth}{!}{%
\begin{tabular}{|l|c|r|c|}
\hline
\textbf{Model}  & \textbf{Type} & \textbf{Size}&\textbf{Layers} \\
huggingface repo used&&&\\
\hline
\textbf{Qwen3-32B}, \citep{qwen3} & Dense & 32B tot& 64\\
Qwen/Qwen3-32B 9216db5&&&\\
\hline
\textbf{Qwen3-235B}, \citep{qwen3}&MoE&235B tot &94\\
Qwen/Qwen3-235B-A22B-Thinking-2507-FP8 f07f63f&&22B act&\\
\hline
\textbf{gpt-oss-120b}, \citep{openai}& MoE & 117B tot & 36 \\
openai/gpt-oss-120b b5c939d& & 5B act & \\
\hline
\textbf{Llama-3.3}, \citep{llama3} & Dense & 70B tot& 80\\
unsloth/Llama-3.3-70B-Instruct (unsloth mirror) 99cd0d2&&&\\
\hline
\textbf{Mixtral-8x22B}, \citep{mixtral8x22b}, \citep{mixtral}, &MoE&141B tot&56\\
mistralai/Mixtral-8x22B-Instruct-v0.1 cc88a6c&&39B act&\\
\hline
\textbf{Mistral Large 2}, \citep{mist}&Dense&123B tot&88\\
mistralai/Mistral-Large-Instruct-2411 ba78820&&&\\
\hline
\textbf{GLM-4.5-Air}, \citep{air}&MoE&106B tot&46\\
zai-org/GLM-4.5-Air a24ceef&&12B act&\\
\hline
\end{tabular}%
}
\caption{Models used in this work}
\label{tab:models}
\end{table}
Table~\ref{tab:models} lists the open-source models used for most of tests in this work. Weights were loaded with transformers via AutoModelForCausalLM.from\_pretrained(repo, revision=\texttt{<}commit\texttt{>}). Each revision was the repository's main HEAD at access time (July 2026).

\section{Objects}
\label{app:obj}
Our test set consists of 3 groups. The first contains the 85 stars of apparent magnitude 2.5 or brighter under their proper names, which, together with their coordinates, are taken from IAU WGSN catalog. The second group has 15 miscellaneous, arbitrarily chosen night sky objects like prominent galaxies, nebula and clusters (see Table~\ref{tab:misc}), whose coordinates are pulled via astropy's SkyCoord.from\_name method. 

\begin{table}[h]
\centering

\begin{tabular}{|r|r|r|}
\hline
\textbf{GALAXIES} & \textbf{NEBULAS} & \textbf{CLUSTERS}\\
\hline
Andromeda Galaxy &	Orion Nebula & Pleiades\\
Triangulum Galaxy &	Crab Nebula & Hercules Cluster\\
Whirlpool Galaxy & Ring Nebula & Omega Centauri\\
Sombrero Galaxy	& Lagoon Nebula & Beehive Cluster\\
Large Magellanic Cloud & Carina Nebula &\\
& Tarantula Nebula&\\
\hline
\end{tabular}%

\caption{Miscellaneous objects included in this study}
\label{tab:misc}
\end{table}

Lastly, we often include 88 official constellations. For purposes of our analysis we define and compute the ``location'' of constellations on the night sky as their area centroids: $\sim 2\cdot 10^5$ points are sampled uniformly over the unit sphere, those that fall within the constellation (determined by astropy's get\_constellation) are averaged, and the result is projected back onto the unit sphere. We have two arguments to excuse the arbitrary nature of this choice. The first is that successful models in our tests don't perform worse on constellations compared to objects with less subjective point representatives (this analysis is not included), so the data couldn't have been spoiled too much by our choice of designation. The second argument is that in our tests we got median angular error to be $12^\circ-21^\circ$ at best, so this noise would likely hide any imperfections of our choice anyway, given that the typical radius of a constellation is $10^\circ-15^\circ$, so any reasonable ways to pick a center would differ by a much smaller amount.

\section{Prompts}
\label{app:prompts}

The prompts we used for most tests are listed in  Table~\ref{tab:prompts}. The set of prompts was generated by a combination of composing, weeding out and adjusting by hand, and using Claude to expand existing sets. Originally, the prompts were more diverse, but the better performance was shown if
\begin{itemize}
    \itemsep-.75cm 
    \item the prompt asked about the nearest vicinity (not more general geometric relations)\\
    \item it made clear that it talks about specifically the night, in some way.
\end{itemize}
Thus, only prompts of this type made the final list.
\begin{table}[h]
\centering
\begin{tabular}{|lll|}
\hline
in the night sky X is not far from & on the celestial sphere X is near &sky atlases place X close to\\
in the night sky X shares space with & in star atlases X lies close to &on sky maps X appears beside\\
in the night sky X is a short hop from& in the night sky X appears near &in sky atlases X neighbors\\
in the sky charts X is right next to&on star charts X sits beside&on star maps X lies beside\\
in the night sky X is close beside&sky watchers locate X next to&sky charts show X beside\\
astrophotographers find X by first locating&in sky atlases X is adjacent to&nearby in the sky to X is\\
in the night sky X is right next to&the brightest star near X is &in the night sky X borders\\
the neighboring constellation to X is&stargazers spot X right beside&sky maps place X near\\
the closest constellation to X is & &\\
\hline
\end{tabular}%

\caption{Default list of prompt templates}
\label{tab:prompts}
\end{table}

To show the significance of prompt phrasing being focused on night sky proximity, we run the same test on a more generic set of prompts listed in Table~\ref{tab:noloc_prompts}. The results (Appendix~\ref{app:noloc}) show that, in this case, the representation appears weaker, never touches the top-8 PCA components (despite residing there entirely under default prompts), and decays quicker. 

\begin{table}[h]
\centering
\begin{tabular}{|llll|}
\hline
the age of X is estimated to be&X is&
X was cataloged by&the magnitude of X is\\
the discovery of X is due to&X emits&
X is classified as&X is composed mainly of\\
in mythology, X represents&X is known for&
X is described in&the temperature of X is\\
X was first recognized by&X is studied by&
X is mentioned by&the luminosity of X is\\
the name of X comes from&X is believed to&
brightness of X is&astronomers use X for\\
last time X was seen in&X was named after&
X is an example of&X is important for\\
X was first observed by&&&\\
\hline
\end{tabular}%

\caption{Generic prompts}
\label{tab:noloc_prompts}
\end{table}

\section{Charts}
\label{app:pca}

Here we explain the analysis summarized in Figures~\ref{fig:mist&lama} and~\ref{fig:mist_noloc} and provide them for other models. All $25\times 188$ prompts are fed into the model in the auto-completion (no thinking) regime and for each layer we record the top 128 PCA components of the residual stream on the last token of the prompt. 

\subsection{Plots} The presence of the sky sphere representation is validated with the following leave-one-out (LOO) analysis. The model prediction for an object $X$ is set via linear regression (using all objects but $X$) of top-4 or top-8 PCA space onto the ``true'' night sky globe space --- $\mathbb{R}^3$ where all objects, are placed on the unit sphere $S^2\subset \mathbb{R}^3$ according to their true celestial coordinates. The accuracy of that prediction is measured with $R^2$-score (variance explained by the model) and median angular error $\alpha$.

\begin{figure}[htp]

\begin{subfigure}{0.5\textwidth}
\includegraphics[width=0.95\textwidth]{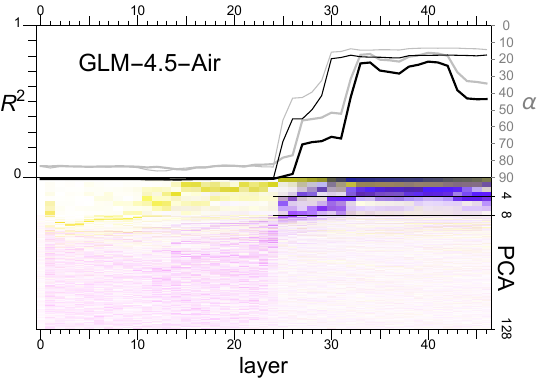}
\end{subfigure}
\begin{subfigure}{0.5\textwidth}
\includegraphics[width=0.95\textwidth]{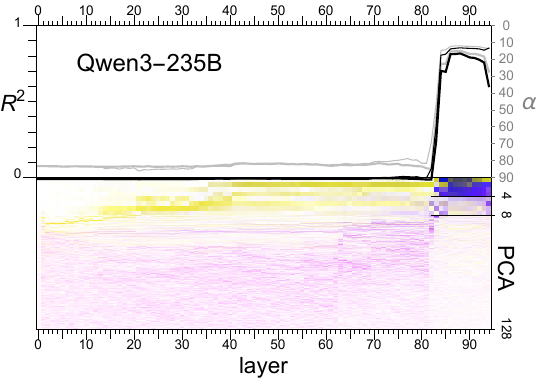}
\end{subfigure}

\bigskip

\begin{subfigure}{0.5\textwidth}
\includegraphics[width=0.95\textwidth]{mistrallarge123b_dump.pdf}
\end{subfigure}
\begin{subfigure}{0.5\textwidth}
\includegraphics[width=0.95\textwidth]{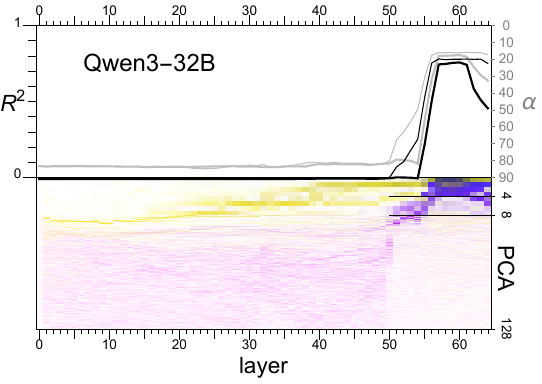}
\end{subfigure}

\bigskip

\begin{subfigure}{0.5\textwidth}
\includegraphics[width=0.95\textwidth]{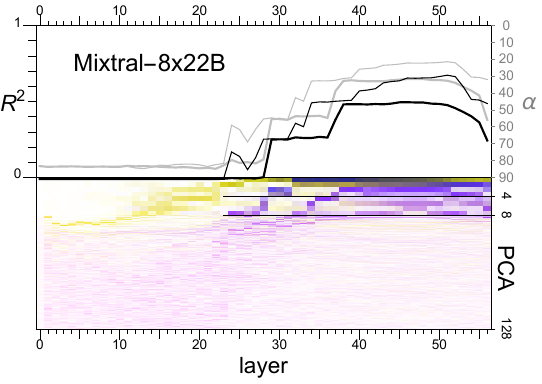}
\end{subfigure}
\begin{subfigure}{0.5\textwidth}
\includegraphics[width=0.95\textwidth]{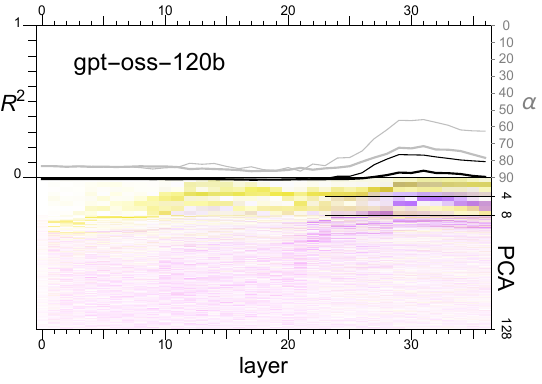}
\end{subfigure}

\bigskip

\begin{subfigure}{0.5\textwidth}
\includegraphics[width=0.95\textwidth]{llama33_70b_dump.pdf}
\end{subfigure}
\begin{subfigure}{0.5\textwidth}
\includegraphics[width=0.95\textwidth]{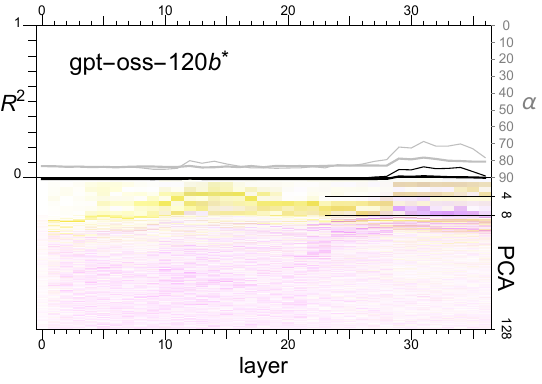}
\end{subfigure}

\caption{$R^2$-score (black) and median angular error (gray) for top-8 (thin) and top-4 (thick) PCA components for a sky sphere fit. Colored charts show which PC's correlate with the optimal projection (blue\&purple) and with the object type (yellow) (see details in text Appendix~\ref{app:pca}). The last gpt-oss-120b$^*$ chart for a special test for this outlier model, see Appendix~\ref{app:gpt}}
\label{fig:dumps}
\end{figure}

\subsection{Density PCA charts} Below the plots we show charts that illustrate the location of the sky sphere representation within PCA directions (blue and purple). Specifically, we center and whiten the sky data (so that their coordinates $c_k$ are distributed with mean 0 and unit variance $I_3$), and record 
$$s(l,i)=\sum_{1\leq k\leq 3}\Cov^2(c_k,(wp)_{li})$$  
where $(wp)_{li}$ is the $i$-th principal component $p$ of $l$-th layer scaled by a weight $w$ that either uniformly scales the layer so that the top variance is 1, or whitens the data so that every component has unit variance. The first version sets the red channel and the second sets the green via formula $1-\sqrt{s(l,i)}$. The blue channel is defined like the green one (with whitened PCA), but instead of correlation with celestial coordinates $c_k$ we correlate $(wp)_{li}$ to the ``type'' function that is 1 if the object $X$ in the prompt is a star, -1 if $X$ is a constellation, and 0 if $X$ is neither. That means that the colors on the chart imply the following:

\begin{table}[h]
\centering

\begin{tabular}{|l|l|l|}
\hline
    Color   & Correlation with & PCA strength\\
\hline \hline
    Dark blue & Sky Sphere, Strong & Strong \\
\hline
    Light blue & Sky Sphere, Weak & Strong\\
\hline
    Dark purple & Sky Sphere, Strong & Weak\\
\hline
    Light purple & Sky Sphere, Weak & Weak\\
\hline
    Solid Yellow & Object Type, Strong&\\
\hline
    Light yellow & Object Type, Strong&\\
\hline
    White & Nothing&\\
\hline
\end{tabular}%

\caption{Meaning  of colors in charts like Figure~\ref{fig:dumps}}
\label{tab:colors}
\end{table}

\subsection{gpt-oss Anomaly}
\label{app:gpt}
\begin{figure}[h]

\begin{subfigure}{0.58\textwidth}
\includegraphics[width=0.75\textwidth]{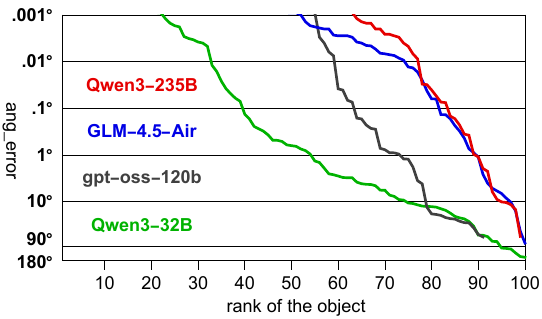}
\caption{Accuracy on prompts ``Give the right ascension and declination of X with no commentary.'', log of angular error sorted in increasing order (results closer than $.001^\circ$ not shown). Some curves don't span the whole 100 tested non-constellation objects since on some obscure stars models have ran out of thinking budget, eliminating them from the pool.}
\label{fig:actual}
\end{subfigure}
\hspace{1cm}
\begin{subfigure}{0.38\textwidth}
\includegraphics[width=0.99\textwidth]{gptoss120b_harmony_dump.pdf}
\caption{Chart from Figure~\ref{fig:dumps} for gpt-oss on prompts with proper harmony response format.}

\label{fig:harmony}
\end{subfigure}
\caption{}
\end{figure}

Figures~\ref{fig:all_r2_ang} and~\ref{fig:dumps} show that in gpt-oss-120b the sky sphere is much less decodable than in models of comparable size. We note that it doesn't mean that ``it is way worse at knowing celestial coordinates''; in fact, all models with thinking in our test managed to locate most non-constellation objects to within~$1^\circ$ (Figure~\ref{fig:actual}) --- way better than $10-20^\circ$ error we get in our tests.

This leaves many questions regarding the role of the representation we found in model's thinking, but also leaves its absence in gpt-oss unexplained. One way in which it differs from other models is that it was trained on the harmony response format (see~\citep{harmony}), rather than raw text, so our raw prompts were out of distribution in that regard. However, when we wrapped prompts in the appropriate format, the decodability representation dropped even lower (Figure~\ref{fig:harmony}, same as last chart in Figure~\ref{fig:dumps})

\subsection{Charts for proximity-focused vs generic prompts}
\label{app:noloc}
Here we illustrate the effect that the phrasing of the prompts has on decodability of sky sphere representation. Note that the apples-to-apples comparison is to be done between the gray shade graph (\hyperref[tab:prompts]{sky proximity} prompts) and thick black line (\hyperref[tab:noloc_prompts]{generic} prompts), both being restricted to the same top-8 PCA space. In this case the signal for generic prompts stays essentially at 0. The $R^2$-score only rises to the thin black line when the domain is expanded to top-32 PCA space, and even then it is far from the results on the focused prompts, and decays in later levels.

\begin{figure}[htp]

\begin{subfigure}{0.5\textwidth}
\includegraphics[width=0.95\textwidth]{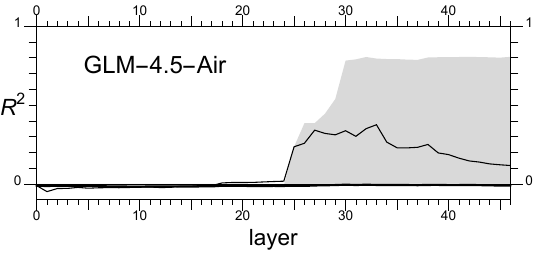}
\end{subfigure}
\begin{subfigure}{0.5\textwidth}
\includegraphics[width=0.95\textwidth]{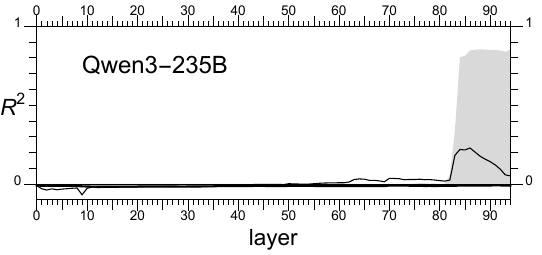}
\end{subfigure}

\bigskip

\begin{subfigure}{0.5\textwidth}
\includegraphics[width=0.95\textwidth]{mistrallarge123b_noloc.pdf}
\end{subfigure}
\begin{subfigure}{0.5\textwidth}
\includegraphics[width=0.95\textwidth]{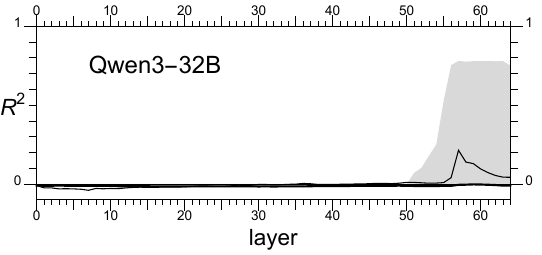}
\end{subfigure}

\bigskip

\begin{subfigure}{0.5\textwidth}
\includegraphics[width=0.95\textwidth]{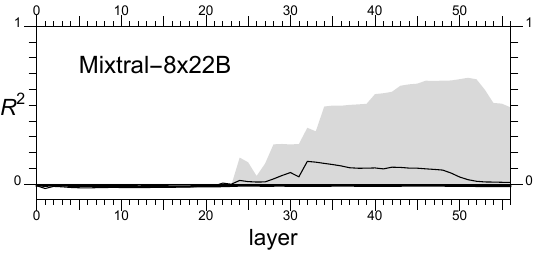}
\end{subfigure}
\begin{subfigure}{0.5\textwidth}
\includegraphics[width=0.95\textwidth]{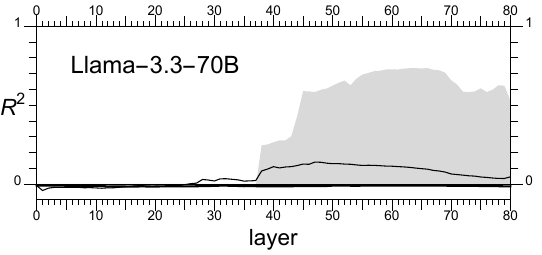}
\end{subfigure}

\bigskip

\begin{subfigure}{0.5\textwidth}
\includegraphics[width=0.95\textwidth]{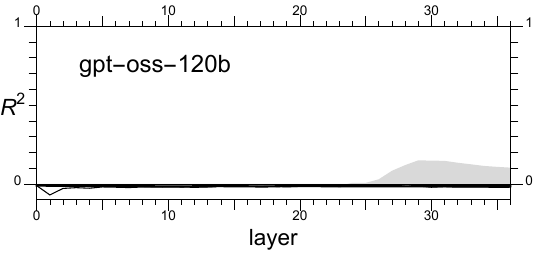}
\end{subfigure}

\caption{$R^2$ LOO scores for decodability of position on the sky sphere from residual stream of prompts focused on \protect\hyperref[tab:prompts]{sky proximity} (gray shade, top-8 PCA only) and \protect\hyperref[tab:noloc_prompts]{generic} prompts (thick black - on top-8 PCA components, almost 0; thin black - on top-32 PCA components).}
\label{fig:noloc}
\end{figure}

\newpage
\section{Testing alternative representations}
\label{app:other_maps}
We argue that the representation we found is indeed based on a spherical feature manifold, rather than a leak from a conceptually different but correlated representation. To that end, we pit our model against two other possible representations of the night sky objects. We present evidence that they are essentially absent from the top-8 PCA where the spherical representation resides.

\subsection{Flat chart}
\label{app:other_maps_flat}
Since the location on the night sky is typically documented as RA/Dec\footnote{Right Ascension and Declination.} coordinates, that are night sky equivalents of longitude and latitude, one might expect a representation that is linear in those coordinates (a flat 2-D map). This is the representation analyzed in~\citep{spacetime} for the case of continents on the globe and states of USA.

\begin{figure}[htp]

\begin{subfigure}{0.5\textwidth}
\includegraphics[width=0.95\textwidth]{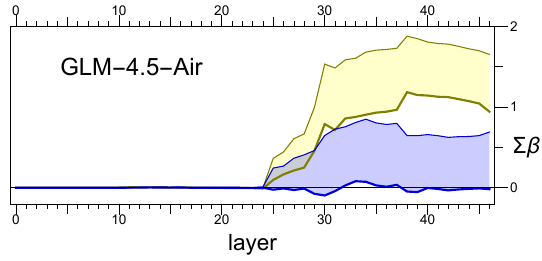}
\end{subfigure}
\begin{subfigure}{0.5\textwidth}
\includegraphics[width=0.95\textwidth]{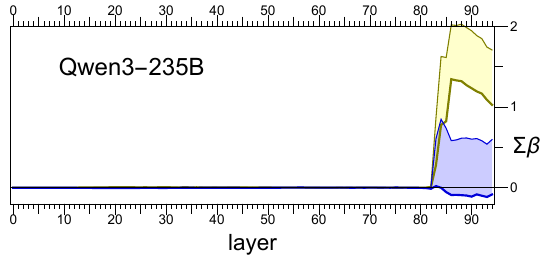}
\end{subfigure}

\bigskip

\begin{subfigure}{0.5\textwidth}
\includegraphics[width=0.95\textwidth]{mistrallarge123b_vs_models.pdf}
\end{subfigure}
\begin{subfigure}{0.5\textwidth}
\includegraphics[width=0.95\textwidth]{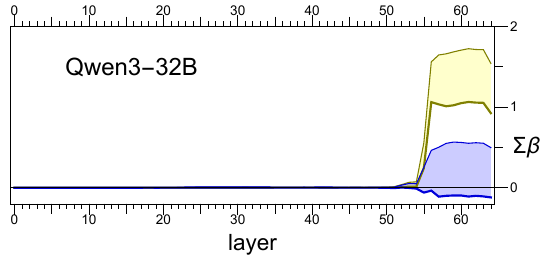}
\end{subfigure}

\bigskip

\begin{subfigure}{0.5\textwidth}
\includegraphics[width=0.95\textwidth]{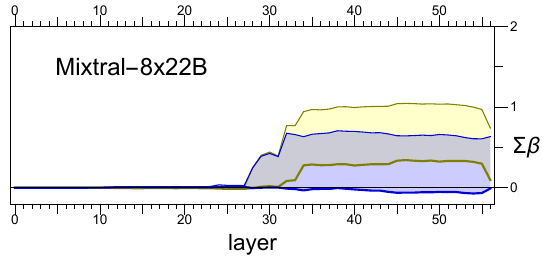}
\end{subfigure}
\begin{subfigure}{0.5\textwidth}
\includegraphics[width=0.95\textwidth]{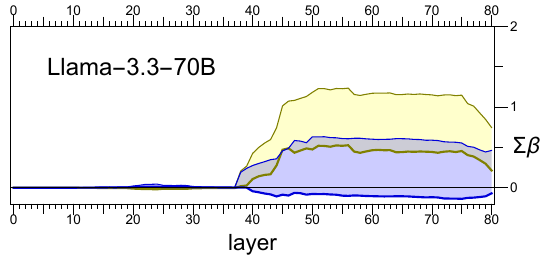}
\end{subfigure}

\bigskip

\begin{subfigure}{0.5\textwidth}
\includegraphics[width=0.95\textwidth]{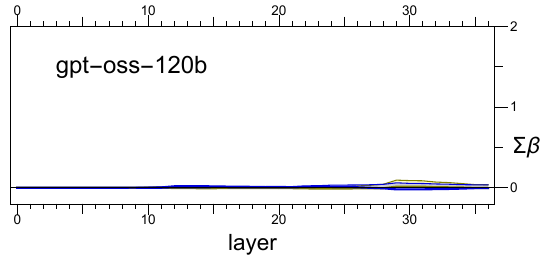}
\end{subfigure}

\caption{Prominence (sums of $\beta$'s in equation~\ref{eq:regression}, $\Sigma\beta$) of the spherical (yellow) and flat 2D (blue) sky representations in top-8 PCA space of the residual stream, either excluding both $\beta_x$ and $\beta_{\Dec}$ (thick bottom lines) or including them both (thin top lines).}
\label{fig:ra}
\end{figure}

It is not trivial to separate this model from the spherical one, since they are highly correlated\footnote{$\Dec$ is almost identical to $x=\sin{(\Dec)}$ of the sphere and $\RA$ is correlated with $y=\sin(\RA)\cos(\Dec)$ of the sphere.}. To distinguish them, we use RDM analysis (compare the distances imposed by each model to those observed in the data). In short, we see that the RA-coordinate (the most distinctive feature of RA/Dec representation) is absent in top-8 PCA (bottom blue line in Figure~\ref{fig:ra} is $\leq0$).

Here are the details. We exclude from the pool the constellations that cross the $\RA=0^\circ\equiv360^\circ$ discontinuity. For each layer $l$ we take top-8 PCA space of the residual stream\footnote{Fitting linearly the RA/Dec model mostly put it aligned with the sphere representation at the top, so we tested whether RA/Dec shares the space with it in the top rather than sitting separately in lower components.}, scale it to make the top variance be 1, and compute the RDM matrix $M$ defined via 
$$M_{ij}=|v_i-v_j|^2$$
where $v_i$ is the vector representing object $i$ (averaged over prompt templates). We use the square of the distance so that such RDM would be additive for direct sums of representations, which we rely on. We perform a linear regression
\begin{equation}
M=\bigl(\beta_xM_x+\beta_yM_y+\beta_zM_z\bigr)+\bigl(\beta_{\RA}M_{\RA}+\beta_{\Dec}M_{\Dec}\bigr)+\eta
\label{eq:regression}
\end{equation}
where $\eta$ is the minimized error, $M_\chi$ is the RDM matrix defined similarly for one of the \hyperref[tab:coords]{coordinates} (normalized to variance 1), and $\beta_\chi$ is the corresponding parameter.

\begin{table}[h]
\centering
\resizebox{\columnwidth}{!}{%
\begin{tabular}{|c|c|}
\hline
Sphere model & 2D flat chart\\
\hline
$x=\sin(\Dec)$\hspace{1cm} $y=\cos(\Dec)\sin(\RA)$\hspace{1cm} $z=\cos(\Dec)\cos(\RA)$& RA\hspace{.5cm} Dec\\
\hline
\end{tabular}%
}
\caption{Coordinate components of tested models}
\label{tab:coords}
\end{table}

Since $x$ and $\Dec$ are almost identical, the split between $\beta_x$ and $\beta_{\Dec}$ is numerically unstable and meaningless in practice. To circumvent this issue, we only use them merged as $\beta_{\widetilde{x\Dec}}:=\beta_x +\beta_{\Dec}$ and record the total $\beta$ score for each model with and without this merged term. The results (Figure~\ref{fig:ra}) clearly show that without the inseparable $\beta_{\widetilde{x\Dec}}$, the flat model is more than non-existent: $\beta_{\RA}\leq 0$ (lower thick blue line). On the other hand, the spherical model is prominent even without its $\beta_{\Dec}$ component (thick yellow line). This shows that regardless of which model we attribute the ambiguous altitude component $\widetilde{x\Dec}$, the flat 2D representation cannot be present in the data, since otherwise we would see its unique feature --- the discontinuous latitude $\RA$, --- and we do not.    

\newpage

\subsection{3D chart, radial analysis}
\label{app:other_maps_rad}
Another conceivable alternative representation is something akin to a genuine 3D map of the celestial objects that properly accounts for the distance to them rather than projecting them onto a unit sphere. However, expecting a literal undisturbed 3D map in this case is unreasonable, since the distances to the stars in our set span $\sim3$ orders of magnitude, including galaxies adds several more, and constellations just don't have a meaningful notion of distance to them. Even so, it raises the question of whether the radial direction in the observed representation does encode the distance to the object in some warped way. And if not --- whether it encodes or is influenced by some other factors.

\begin{figure}[htp]
\centering
\includegraphics[width=0.90\textwidth]{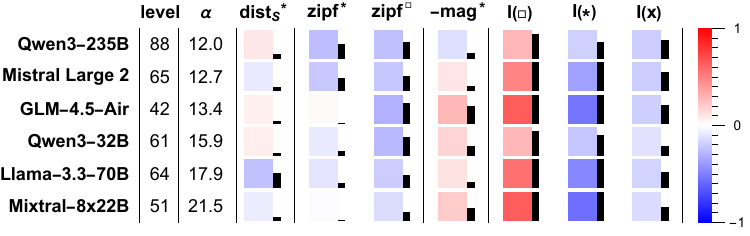}
\caption{Correlations of radial direction with various functions (see Table~\ref{tab:corr_funcs}) at the \textbf{level} with least median angular error $\alpha$. Black bars represent significance of the correlation, height is
 $1-p^{\ln(.5)/\ln(.05)}$ ($p=0.05$ gives height 0.5).}
\label{fig:radcor}
\end{figure}

Figure~\ref{fig:radcor} shows the correlations of the radial direction with various parameters, with black bars indicating significance of the correlation ($p=0.05$ corresponds to half-height bar). We see that (Spearman) correlations with distance ($\mbox{\textbf{dist}}_S^*$) are weak and disparate with low significance (only Llama barely climbs past $p=0.05$). Given the textual origin of the data, the textual frequency (\textbf{zipf}) is the next factor that we test against, with more pronounced results, especially for constellations ($\mbox{\textbf{zipf}}^\square$) rather than stars ($\mbox{\textbf{zipf}}^*$), placing more mentioned objects slightly closer. Brightness is a parameter that is both physical and reflecting how noticeable it is for people. We see mostly positive, but weak and low significance correlations of radial direction with brightness (\textbf{-mag}, negative apparent magnitude). The only really robust correlations we found were with the object type, placing constellations farther ($\mbox{\textbf{I}}(\square)>0$) and stars closer ($\mbox{\textbf{I}}(*)>0$).

To do this analysis, we pick the level with the best median angular error $\alpha$ in LOO analysis on top-8 PCA subspace (computed as in Appendix~\ref{app:pca}), and discard from the pool the objects whose MSE for the angle is more than double of the average, to eliminate outliers and get a cleaner signal. We use linear regression to project the top-8 PCA space onto the 3-dimensional space that hosts the unit sphere sky model. For the remaining objects we compute correlations of the length of their projection with the following functions listed in Table~\ref{tab:corr_funcs}.

\begin{table}[h]
\centering

\begin{tabular}{|c|l|r|}
\hline
    Label   & Function correlated & Domain\\
\hline \hline
    $\mbox{dist}_S^*$ & Distance, from SIMBAD trigonometric parallax; Spearman correlation & Stars only \\
\hline
    $\mbox{zipf}^*$ & Log frequency in text corpus, from wordfreq.zipf\_frequency & Stars only\\
\hline
    $\mbox{zipf}^\square$ & Log frequency\dots, `` constellation'' stripped &Constellations only\\
\hline
    $-\mbox{mag}^*$ & Brightness as negative apparent magnitude, from SIMBAD & Stars only\\
\hline
    $I(\square)$ & Indicator function of ``constellation'' type& All\\
\hline
    $I(*)$ & Indicator function of ``star'' type& All\\
\hline
    $I(\mbox{x})$ & Indicator function of ``other'' type& All\\
\hline
\hline
\end{tabular}%

\caption{Functions correlated with radial length.}
\label{tab:corr_funcs}
\end{table}

\newpage

\section{PCA components ahead of sky sphere}
\label{app:hogs}

\begin{figure}[htp]
\centering
\includegraphics[width=0.90\textwidth]{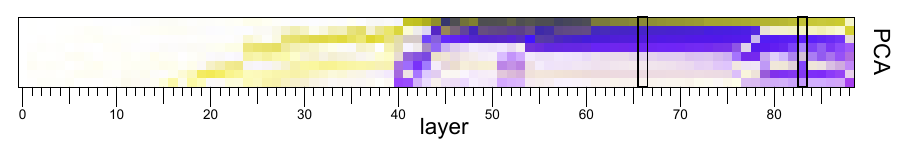}
\caption{Partial copy of Mistral chart from Figure~\ref{fig:dumps} with analyzed layers 66 and 83 marked.}
\label{fig:mist_marked}
\end{figure}

In many models the sky sphere representation rises to the very top of PCA components (blue in Figure~\ref{fig:dumps}) on proximity focused prompts. However, even in the best examples it falls short from crystallizing as the top-3 components (see for example top-8 PCA of Mistral model, Figure~\ref{fig:mist_marked}). This is mostly due to the direction (yellow) that separates concepts ``star'' and ``constellation'' prevalent in our test data (see Appendix~\ref{app:pca} for the definition of the yellow direction).

\begin{figure}[htp]

\begin{subfigure}{0.4\textwidth}
\includegraphics[width=0.95\textwidth]{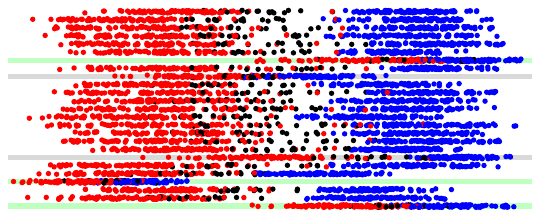}
\caption{Mistral Level 66 top-4 PCA components minus the sky sphere:\newline \hspace{1cm} object is \textbf{\textcolor{red}{STAR}}, \textbf{\textcolor{blue}{CONSTELLATION}}, \textbf{\textcolor{black}{NEITHER}}}
\label{fig:l66_PCA(4-1)}
\end{subfigure}
\begin{subfigure}{0.6\textwidth}
\includegraphics[width=0.95\textwidth]{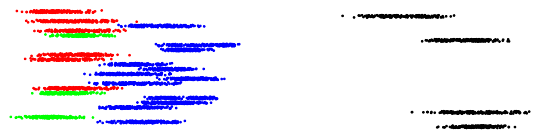}
\caption{Mistral Level 83 top-2 PCA components minus the ``type'' direction:\newline \hspace{1cm} last word is \textbf{\textcolor{black}{IS}}, \textbf{\textcolor{red}{BESIDE}}, \textbf{\textcolor{green}{NEAR}}, \textbf{\textcolor{blue}{other}}}

\label{fig:l83_PCA(2-1)}
\end{subfigure}

\caption{Top PCA directions ahead of the sky representation in Mistral Large 2.}

\end{figure}

Another way to analyze the competitor direction is to pick the top-4 components for a layer, where they are saturated with the sky sphere model, like level 66 in Mistral model, and isolate the direction orthogonal to the sky model (\cref{fig:l66_PCA(4-1)}). The prompts whose object $X$ is a constellation (blue) land on one side and those where $X$ is a star (red) land on the other, with miscellaneous objects (black) spread in between. The three notable exceptions (highlighted in green) correspond to the cases where the template itself contains the word ``constellation'' or ``star'', landing them firmly in the corresponding group. Two softer outliers (highlighted in gray) are the templates
\begin{itemize}
    \itemsep-.75cm    
    \item ``in the night sky X \textbf{borders}'', tilts towards constellations,\\
    \item ``\textbf{astro}photographers find X by first locating'', tilts towards stars,\\
\end{itemize}
which is in line with our interpretation: a vague relation to the concept produces a mild bias.

While the ``star''-``constellation'' direction is the only unsurpassed rival to the sky sphere in many models, we can also clearly see another direction surface up in layers 79-87 in Mistral. Plotting the direction, orthogonal to ``star''-``const'' direction in layer 83, reveals its meaning (\cref{fig:l83_PCA(2-1)}): it is focused on the last word of the prompt. 

We note that while we provided the plots only for one layer in each case, the plots for all layers 54-75 and 79-87, respectively, are essentially the same.

\end{document}